\documentclass{article}
\usepackage{PRIMEarxiv}

\usepackage[utf8]{inputenc} % allow utf-8 input
\usepackage[T1]{fontenc}    % use 8-bit T1 fonts
\usepackage{hyperref}       % hyperlinks
\usepackage{url}            % simple URL typesetting
\usepackage{booktabs}       % professional-quality tables
\usepackage{amsfonts}       % blackboard math symbols
\usepackage{nicefrac}       % compact symbols for 1/2, etc.
\usepackage{microtype}      % microtypography
\usepackage{lipsum}
\usepackage{fancyhdr}       % header
\usepackage{graphicx}       % graphics
\usepackage{amsfonts}
\usepackage{amsmath}
\usepackage{graphicx}
\usepackage{qtree}
\usepackage{pict2e}
\usepackage{eepic}
\usepackage{color}
\usepackage{moreverb}
\usepackage{fancyvrb}
\usepackage{tabularx}
\usepackage{colortbl}
\usepackage{acronym}
\usepackage{multirow}
\usepackage{longtable}
\usepackage{paralist}
\usepackage{url}
\usepackage{xspace}
\usepackage{multirow}			% multirow nas tabelas
\usepackage{subcaption}
\usepackage{algorithm}
\usepackage{algorithmicx}
\usepackage{algpseudocode}
\usepackage{verbatim}			% verbatim, comment, etc.
\usepackage{array}				% Extens„o de tabular e array
\usepackage{colortbl}			% Tabelas com cores
\usepackage{tikz-qtree}
\usepackage{rotating}
\usepackage{etoolbox}
\usepackage{indentfirst}
\usepackage{booktabs}
\usepackage{epstopdf}
\usepackage{bm}
\usepackage{natbib}
\usepackage{authblk}

\graphicspath{{media/}}     % organize your images and other figures under media/ folder
\acrodef{BSL}{British Sign Language}
\acrodef{LGP}{Portuguese Sign Language}
\acrodef{FACS}{Facial Action Coding System}
\acrodef{AUs}{Action Units}
\acrodef{ASL}{American Sign Language}
\acrodef{CORS}{Cross-Origin Resource Sharing}
\acrodef{IK}{Inverse Kinematics}
\acrodef{FK}{Forward Kinematics}
\acrodef{FE}{Facial Expressions}
\acrodef{EP}{European Portuguese}

\author[1,2]{Inês Lacerda}
\author[1,3]{Hugo Nicolau}
\author[1,2]{Luísa Coheur}
\affil[1]{Instituto Superior Técnico, Universidade de Lisboa}
\affil[2]{INESC-ID}
\affil[3]{Interactive Technologies Institute / LARSYS}

\title{Enhancing Portuguese Sign Language Animation with Dynamic Timing and Mouthing}

\begin{document}
   \maketitle
\begin{abstract}
 Current signing avatars are often described as unnatural as they cannot accurately reproduce all the subtleties of synchronized body behaviors of a human signer. In this paper, we propose a new dynamic approach for transitions between signs, focusing on mouthing animations for Portuguese Sign Language. Although native signers preferred animations with dynamic transitions, we did not find significant differences in comprehension and perceived naturalness scores. On the other hand, we show that including mouthing behaviors improved comprehension and perceived naturalness for novice sign language learners. Results have implications in computational linguistics, human-computer interaction, and synthetic animation of signing avatars.
\end{abstract}

\section{Introduction}

Spoken/written language and sign language are different: one is an audio-oral language while the other is a spatial-visual language. Vocabulary and grammatical rules are also quite different. These differences lead to a language barrier between Deaf\footnote{Deaf with a capital refers to people who identify with the deaf culture and have been deaf before they started to learn a language. They are pre-lingually deaf.} and hearing people. In this paper, we contribute to breaking down the language barrier between European Portuguese and \ac{LGP}.

Sign language translators typically require two components: a language translator and a signing avatar. The translator converts written text (or speech) into a sequence of glosses (i.e., lexical units representing each gesture or sign); then, the avatar displays the synthesized glosses and additional linguistic components as signing animations. Signing avatars must account not only for multiple co-occurring linguistic processes but also for the naturalness of the movements. However, most avatars are described as unnatural, emotionless, and stiff~\cite{kipp2011assessing} because they cannot accurately reproduce a human signer's subtleties. Therefore, one of the goals of an automatic sign language translator is to perform secondary movements based on
human kinematics, so that animations are understandable and perceived as natural.

Planning and scripting a signing avatar's facial and body movements is difficult. Minor variations in timing and speed parameters can lead to significant differences in the quality and understandability of sign animations~\cite{huenerfauth2009linguistically, al2018modeling}. In the case of sign languages, transitions between signs rely heavily on the phonology of the previous and following signs and determine the movement fluidity that allows sign streams to be intelligible. Therefore, transitions can impact the comprehension and naturalness of sign animations. In this paper, we introduce a new approach for interpolating signs, \textbf{dynamic transitions}, which change according to the previous and following signs. We aimed to answer the following research question:

\textit{RQ1: Do dynamic transitions improve linguistic comprehension, naturalness, and preference of sign Language animations?}

We conducted a user study with 11 native speakers to understand the effect of dynamic transitions and optional transition speed. Overall, results show that dynamic transitions are equivalent to constant transitions in each of the former dimensions (comprehension, naturalness and preference). However, dynamic transitions enhance linguistic comprehension for signs that comprise one sole meaning (i.e., composite utterances and negatives) and require faster transitions. 

\begin{figure} [ht]
    \centering
    \includegraphics[width=7cm]{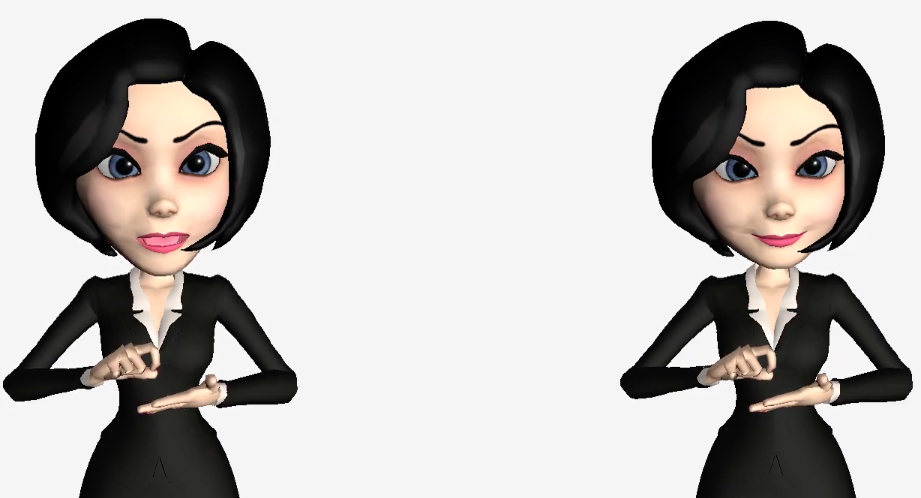}
    \caption{Avatar with (left) and without mouthing (right).}
    \label{fig:mouthing}
\end{figure}

In addition to transitions between signs, the avatar's \textbf{secondary movements} -- those that are added to improve the naturalness of the avatar and are not part of the morphosyntactic structure of sign languages -- can also greatly impact the signing quality and the way the thought or feeling is conveyed. Secondary movements include eye blink, mouthing, and facial and corporal movements. In this paper, we study the effect of adding \textbf{mouthing} -- the production of visual morphemes or syllables that derive from spoken language (Figure \ref{fig:mouthing}) -- to an existing avatar.

Some believe that mouthing is incorporated into the morphosyntactic structures of sign languages, and some believe they are not~\cite{crasborn2008frequency, wolfe2018exploring}. Nevertheless, it is interesting that mouthing can be combined with manual signs to create complex signs with a composite meaning. For example, the manual sign ``mouse'' in \ac{BSL} can be accompanied by the mouth action ``baby'', forming the composite meaning ``baby mouse''~\cite{crasborn2008frequency}. Notice that mouthing should only be reproduced when there are no phonological facial expressions that contain the mouth (e.g., cheeks puffed, tongue touching the chin, morphemes). Thus, our second research question was:

\textit{RQ2: Does mouthing impact linguistic comprehension, naturalness, and preference of sign language animations?}

Results show that the avatar with mouthing achieves better results in terms of comprehension, naturalness, and preference. To the best of our knowledge, research in the field has not yet been published on whether mouthing can improve comprehension; therefore, this user study can give valuable input into this topic. 

%It should also be mentioned that, along all this study, we always had feedback from \ac{LGP} experts.
The paper is organized as follows: in Section \ref{sec:rw} we present related work, in Section \ref{sec:syn} we present our models and in Section \ref{sec:exp} we evaluate them. Then, in Section \ref{sec:conc}, we present the main conclusions and future work.

\section{State of the Art}\label{sec:rw}

Research regarding hand signs and facial expressions in Sign Language animations is scarce, and in a synthetic context, the blending of the two is still an open challenge. Existing solutions rely on Sign Language Annotations~\cite{elliott2000development}, Keyframe Animations~\cite{adamo20083d}, and Motion Capture methods~\cite{gibet2011signcom}. Each approach provides advantages and disadvantages but all require a balance between quality and cost. The more accurate and natural the animations are, the more costly they are to be generated.

Some work regarding modeling timing and pausing parameters for manual components has been done for \ac{ASL}~\cite{huenerfauth2009linguistically, al2018modeling}. However, to the best of our knowledge, no study proposes any kind of dynamic transition that rely on the phonology of the previous and following signs for sign language animations.

Prior research has been dedicated to mouthing. Mouthing or lip-sync appeared first in the 1920s with the advent of sound cartoons \cite{wolfe18}. Speech can be discretized as a sequence of sounds known as phonemes. Each phoneme is associated with a facial pose; however, not all vocal articulations are visible, and some are irrelevant in the visual domain. For instance, nasality and voicing \cite{serra12}. Phonemes usually have many-to-one relationships with visemes (i.e., facial and oral poses of phonemes) because different phonemes can have the same facial pose. It is also important to note that visual speech cannot be directly generated by concatenating visemes because it will over-articulate the produced animation.

Mouthing animations can be produced manually or automatically~\cite{wolfe2018exploring}. A manual approach requires animators to draw each viseme by hand and later use an interpolation scheme that concatenates the visemes according to the animated utterances. This method is time-consuming since it requires animators to manually select the viseme and its timing. These limitations led to the development of automatic techniques that synchronize audio with visemes. In automation approaches, visemes are collections of 3D data and artists can rely on muscle-based systems or blend shapes expressed as polygon meshes to model avatars' lip positions to depict visemes.

Automated techniques depend on the source of dialog to generate animation. If it is a pre-recorded voice track, a speech recognition system can be used; otherwise, if it is a text containing a dialog, a text-to-speech system must be used. Both techniques require the same process: detecting the phonemes and then selecting the corresponding visemes that can be interpolated between keyframes in the avatar. No matter the technique, the best mapping between phonemes and visemes is still a debatable issue. Many studies have been developed to understand the best Phoneme-Viseme mappings. The work presented in~\cite{bear2014phoneme} examined 120 mappings and analyzed their effect on visual lip reading using hidden Markov model (HMM) recognizers. Although phoneme-viseme mappings are not universal among languages or within a language, some phoneme-viseme mappings have overlapping sets. For instance, similarly to English, Amazon Polly's Phoneme-Viseme mapping\footnote{\url{https://docs.aws.amazon.com/polly/latest/dg/ph-table-portuguese.html}} for European Portuguese also contains the viseme [/p/ /b/ /m/] as a set, even though these are two different languages.

Some projects have been exploring the possibility of incorporating mouthings in Sign Language animations. The ViSiCAST project~\cite{elliott2000development} developed the Signing Gesture Markup Language, which is an XML-compliant representation of signs based on HamNoSys\footnote{\url{http://www.sign-lang.uni-hamburg.de/dgs-korpus/index.php/hamnosys-97.html}}. In this project, phonemes are described based on the International Phonetic Alphabet (IPA) transcription and then visemes are mapped using the Speech Assessment Methods Phonetic Alphabet (SAMPA) encoding conventions~\cite{glauert2004virtual}. Some work in the area of visual speech animation has been done for \ac{ASL} and Swiss German Sign Language~\cite{wolfe2018exploring}.

The work described in \cite{serra2012proposal} is the first automatic visual speech system for \ac{EP} based on viseme concatenations. This project used two phoneme-viseme mappings: one mapping with 14 different viseme classes and another mapping with 10 different viseme classes. Both Phoneme-Viseme mappings resulted in slightly different vowel classifications, but the number of vocalic viseme classes remained unchanged. Each viseme class was then created in the avatar by an experienced digital artist. After creating the mappings and the visemes, the system was divided into two main components: a speech processing component and a 3D animation engine. The speech process component processed the data (e.g., text, audio, or both) and obtained the phonetic transcriptions using an \ac{EP} phonetic lexicon developed by Microsoft together with Microsoft Speech API (SAPI)\footnote{\url{https://docs.microsoft.com/en-us/previous-versions/windows/desktop/ms723627(v=vs.85)}} as an automatic speech recognition (ASR) model. The SAPI system used recognition events to detect the different utterances from the audio and stored a list of words and their corresponding IPA formats. The SAPI does not provide the phonemes' duration and timing, therefore, the \ac{EP} phonemes' duration was gathered from a database of 100 hours of Portuguese speech provided by Microsoft. The 3D animation engine encapsulated the data obtained from the speech process component and translated it into the 3D animation. The cartoon character relied on a bone-based rig, and each viseme was interpolated using the timing obtained by the speech process component and the animation curves defined by the animator. 

\begin{figure*}[h!t]
    \centering
    \includegraphics[width=12cm]{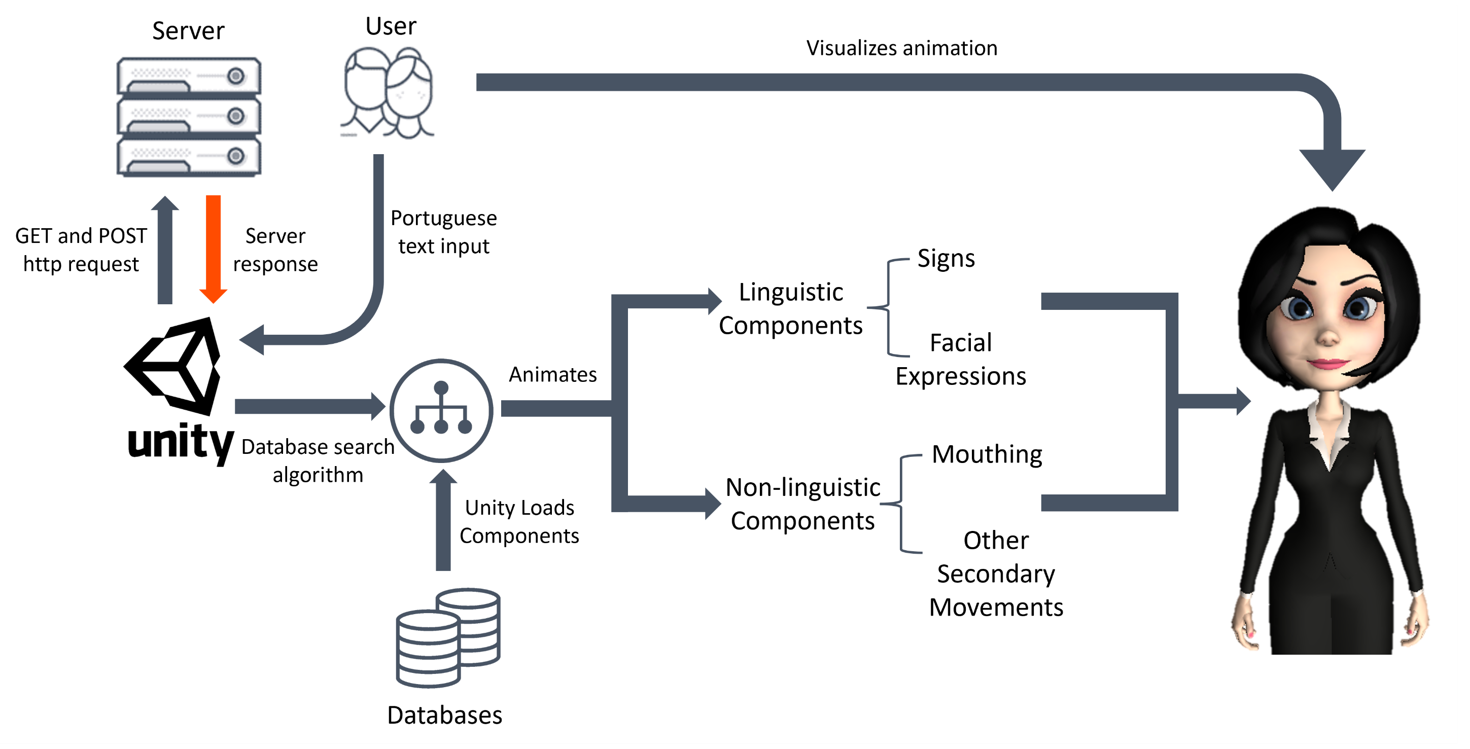}
    \caption{Overall architecture of the text-to-sign translator.}
    \label{fig:communcation}
\end{figure*}

Finally, we highlight the work developed in~\cite{crasborn2008frequency}, where the authors studied the mouth actions from a cross-linguistic perspective for three European Sign Languages: Sign Language of the Netherlands (NGT), British Sign Language (BSL), and Swedish Sign Language (SSL). Considering all the possible mouth action, the authors concluded that mouthing is the mouth action with the highest values, almost in all three languages (57\% in SSL, 51\% in BSL, and 39\% in NGT). In this study, the authors also confirmed the hypothesis that mouthings spread analogously to ``native'' mouth gestures; thus, mouthings have indeed a grammatical function in sign languages. Based on all results gathered from the study, mouthing occurs for all three sign languages, and without it, a signing avatar would look unnatural and could omit important information, thus, resulting in incomprehensible utterances. Therefore, we can conclude that an avatar capable of producing mouthing is an essential part of any automatic written/spoken to sign translation system.
\section{Synthesis of Sign Language Animations}\label{sec:syn}

We used an existing text-to-sign language translator~\cite{pedro, carolina, matilde} to evaluate our animations (overall architecture in Figure \ref{fig:communcation}). This system is divided into two main modules. The first module performs a translation process, consisting of the translation of text from Portuguese into \ac{LGP}, in which the \ac{LGP} sentence is represented by a sequence of glosses and additional morphosyntactic information. The second module consists of an avatar that animates the LGP translated message received from the first module, by using a database with synthesized signs (i.e., animations).

In the following sections, we describe the dynamic transitions and how we have implemented the mouthing process.

\subsection{Dynamic transitions}

%In a first approach, to transition between signs, the state transitions\footnote{\url{https://docs.unity3d.com/Manual/class-Transition.html}} in the animation layers were used, however, their duration and offset values are constant and cannot be changed dynamically in run-time without the use of the \textit{UnityEditor} package. 

Considering that transitions between signs rely heavily on the phonology of the previous and following signs and determine the movement fluidity that allows sign streams to be intelligible, we propose dynamic transitions, which interpolate signs through transitions that change according to the previous/following signs. 

%We use the dictionary created in Section~\ref{chapter:loading_database} that stored the position of the avatar's hands in the first and last keyframes for all signs. 

\begin{comment}
\begin{figure*} [ht]
    \centering
    \includegraphics[width=10cm]{dynamic_transitions_equation}
    \caption{Code to calculate the Dynamic Transitions.}
    \label{fig:dynamic_transitions}
\end{figure*}
\end{comment}
%as shown in Step 1 in Figure \ref{fig:dynamic_transitions}. 
\begin{comment}
 (Equation \ref{eq1}).
 
 \begin{equation}
\begin{split}
RightHandPosDiff =
(CurrentSign[firstFrame] - \\ perviousSign[lastFrame]).sqrMagnitude\\
LeftHandPosDiff =
(CurrentSign[firstFrame] - \\ perviousSign[lastFrame]).sqrMagnitude
\end{split}
\label{eq1}
\end{equation}
\end{comment}

While we iterate over each gloss in run-time, the differences between hand positions in the last keyframe of the previous sign and the first keyframe of the following sign are calculated and then the squared magnitude of these vectors is computed\footnote{The hand positions are obtained from the bones of the skeleton. While creating the database that contains the synthesized sign animations, we also created a JSON file that contains, for each sign and in each keyframe, the facial expressions used, and the hand positions based on the bones of the skeleton. In runtime, we use this information to obtain the hand positions to generate the dynamic transitions.}. Calculating the squared magnitude of a vector is much faster than using the magnitude property since it does not require a slow square root operation that makes the magnitude property take longer to execute\footnote{\url{https://docs.unity3d.com/ScriptReference/Vector3-sqrMagnitude.html}}. %%%%

These squared magnitude values are then converted to percentages by defining a scale. To decide this scale, we checked all signs created to find two signs that have the closest hand position differences (e.g., signs ``EU'' and ``TER'') and two signs that have the furthest hand position differences (e.g., signs ``ELE'' and ``TER''). Based on our findings, we defined two scales: one that includes both hands (if the left hand has movement), and another that only considers the right hand (if the left hand has no movement). Using these scales, the squared magnitude values are converted to percentages that range between 0\% and 100\%. 
\begin{comment}
(Equation \ref{eq2}).
%, as shown in Steps 2 and 3 in Figure \ref{fig:dynamic_transitions}. 

\begin{equation}
\begin{split}
if (BothHands) \\
percentage = (RightHandPosDiff + LeftHandPosDiff) - 0.01
(CurrentSign[firstFrame] - \\ perviousSign[lastFrame]).sqrMagnitude\\
LeftHandPosDiff =
(CurrentSign[firstFrame] - \\ perviousSign[lastFrame]).sqrMagnitude
\end{split}
\label{eq1}
\end{equation}
\end{comment}

Finally, to find the duration value used in the transition between signs, we use the percentage calculated to linearly interpolate between two duration values. These two duration values correspond to the lowest and highest values that the duration of transitions can take. We defined these values by analyzing the lowest and highest transition duration in multiple videos of an \ac{LGP} corpus\footnote{\url{https://portallgp.ics.lisboa.ucp.pt/corpus_lgp/}}. Furthermore, two empirical studies developed by Sedeeq~\cite{al2020empirical, sedeeq2021assets} found that ASL signers prefer slower transitions than the timing of human signers and that they prefer animations with an average transition time of 0.5 seconds. Based on the analysis of our \ac{LGP} corpus, we decided that the duration of transitions would range between 0.3 seconds and 1.1 seconds because this range would include 0.5 seconds as the average transition time and these are slightly slower than the human signing transitions in our \ac{LGP} corpus.

Using the calculated duration values in the process previously described, we created an interpolation between the current sign and the next sign using dynamic transitions by defining a duration value and an offset value\footnote{The offset was introduced as a workaround due to the limitations of the Unity animation engine because the animation blending of Unity requires overlap between the animations. Another way to do it would be by using the UnityEditor package which provides much more animation resources to manipulate the avatar’s animation in runtime. However, these scripts cannot be included in the WebGL build which we needed because we want to have the translator deployed in a website.}. The first keyframe of every sign in the database starts at 1 second, which is what allows transitions between signs to be executed without cutting the signs shorter because without it the transition would overlap the beginning of each sign. Using the offset value, we can adjust the timing until the first keyframe matches the transition duration time; therefore, the offset value is 1 second minus the transition duration value. Transitions must be seen as a continuous stream of motion without being too paused because co-articulation, similarly to oral languages, also constitutes an important part of sign languages. To create transitions that are fluid and not too paused between signs, we decided to define the offset value as 1.2 seconds minus the transition value rather than 1 second. Thus, signs will be slightly overlapped and transitions more fluid.

Another aspect taken into consideration was the phonological assimilation processes of composite utterances. Composite utterances are utterances that have meanings derived from the composition of multiple signs (e.g., ``VERMELHO'' + ``MELÃO'' means ``MELÂNCIA'' (``RED'' + ``MELON'' means ``WATERMELON''). Since multiple signs can be combined for one sole meaning, the transitions between these must be smaller than transitions between signs that have separate meanings. This is another reason why dynamic transitions are so important. These can have an impact on the perception of composite utterances if the phonological assimilation processes are not taken into consideration. Based on empirical experiments, we defined 0.2 seconds as the transition duration in-between all signs that comprise a composite utterance. Using the indices that define composite utterances obtained from the translation process, we transition between signs that comprise composite utterances with a transition value of 0.2 seconds, making the transitions for composite utterances faster than transitions for other signs. 

%%%%%%%%%%%%%%%%%%%%%%%%%%%%%%%%%%%%%%%%%%%%%
%%%%%%%%%%%%%%%%%%%%%%%%%%%%%%%%%%%%%%%%%%%%%
%%%%%%%%%%%%%%%%%%%%%%%%%%%%%%%%%%%%%%%%%%%%%
%%%%%%%%%%%%%%%%%%%%%%%%%%%%%%%%%%%%%%%%%%%%%

\subsection{Mouthing}

\begin{figure*} [htbp]
    \centering
    \includegraphics[height=4.0cm]{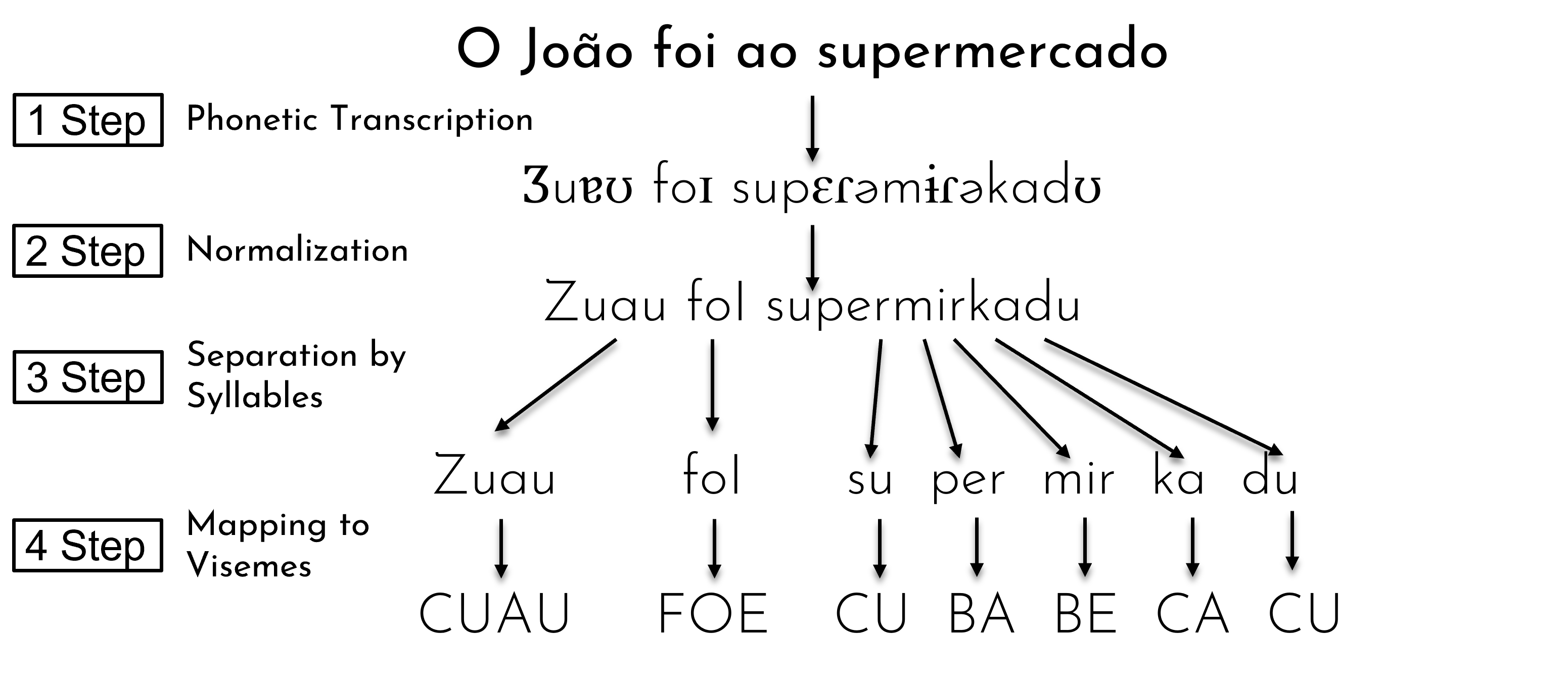}
    \caption{The process of transcribing a sentence phonetically and mapping it to the corresponding visemes.}
    \label{fig:mouthing_process}
\end{figure*}

Mouthing is an essential part of any automatic written-to-sign translation system and without it, a signing avatar would look unnatural and could omit important information. The existing translator was extended to create mouthing animations. Notice that mouthing should be done with words in Portuguese and not their glosses. For instance, verbs are not conjugated while signing, but these should be conjugated while mouthing. Therefore, we extended the system to gather all words in Portuguese and, afterwards, combine them into a sentence so that we consider the assimilation between words when executing the phonetic transcription. (Figure \ref{fig:mouthing_process})

The phonetic transcription (Step 1 in Figure \ref{fig:mouthing_process}) is done by employing the phonemizer tool\footnote{\url{https://github.com/bootphon/phonemizer}}, where the speak backend is used to produce phoneme sequences described based on the International Phonetic Alphabet transcription. After this, normalization (Step 2 in Figure \ref{fig:mouthing_process}) is done by encoding non-ASCII to ASCII, words are separated into their corresponding syllables (Step 3 in Figure \ref{fig:mouthing_process}) using syllabification rules and, then, each phoneme is mapped into one viseme (Step 4 in Figure \ref{fig:mouthing_process}) using the phoneme-viseme mapping we created. While mapping visemes, we need to be careful not to over-articulate as it would generate unnatural mouthing animations. We prevented the over-articulation problem by removing visemes that are irrelevant in the visual domain. For instance, we remove visemes that have equal consecutive visemes, and we remove 'C' viseme consonants (i.e., in a visual domain, represent a slight open mouth) that are at the end of a syllable (e.g., 'r' is removed from 'per' syllable in Figure \ref{fig:mouthing_process}).

The avatar contains 7 visemes -- \textit{A}, \textit{B}, \textit{C}, \textit{E}, \textit{F}, \textit{O}, and \textit{U} -- to animate the 33 phonemes of the Portuguese language. To create visemes as close as possible to human visemes, animations for each viseme were created by adjusting the weights of blend shapes. In the translation process, words are translated into phonemes, separated into syllables, and then mapped into visemes. In the animation process, when the manual signs are being animated, mouthing is animated by using an interpolation scheme that concatenates the visemes according to the animated signs. 

The duration value for the mouthing is defined based on the duration of the sign it is applied to and based on the number of syllables for that sign. The reason behind this is that we do not want mouthing to either overlap the duration of a sign or be too slow if the duration of a sign is too large. The synchronization between mouthing and signs is extremely important because studies~\cite{kipp2011assessing} have reported that a mismatch between the duration of signs and their corresponding mouthings can provoke a disturbing oscillation of the user's visual focus from hands to face.

Finally, we would like to highlight that the mouth movement is slightly anticipated with respect to the hands.
\section{Evaluation}\label{sec:exp}

We have designed and conducted two experimental user studies\footnote{Both studies were approved by the Ethics Committee of our university, and, before participating in the studies, each participant was handed an informed consent that they had to sign to participate.}: we analyze the impact of both dynamic transitions and mouthing on: a) linguistic comprehension, b) naturalness, and c) preference of \ac{LGP} animations. In the case of dynamic transitions, speed was also evaluated. All the signing animations were generated using the previously described sign language translator.

\subsection{Evaluating Dynamic Transitions}
\label{sec:transitions}

We recruited 11 participants fluent in \ac{LGP}. Participants had to fill in a questionnaire. In the  first 10 sections of the questionnaire, participants had to visualize a video, write the understood content, and describe whether the sentence contained an error. As we wanted to evaluate the impact transitions could have on the phonology of signs, particularly on the phonological assimilation of composite sentences, the created sentences contained one or more composite sentences. Furthermore, for each sentence, they also had to evaluate the transitions' speed on a 1-5 Likert scale with one as too slow and five as too fast, and evaluate the avatar's naturalness on a 1-5 Likert scale with one as robotic and five as natural. We presented both conditions - dynamic transitions and constant conditions - to all participants in a counterbalanced order. The order of the first ten sections was randomized.

In the next three sections of the questionnaire, participants had to select the preferred video between two side-by-side videos (one with dynamic transitions and one with constant transitions of 0.5 seconds).  To mitigate experimental bias, the video position was randomized. We also asked participants ``How the naturalness of the avatar could be improved'', whether they think ``transitions between signs affect naturalness'' and whether they think ``transitions between signs affect comprehension''. 

\subsubsection{Comprehension}

For each sentence in the questionnaire, we measured the percentage of content understood by calculating the number of glosses correctly described with 100\% as all glosses correctly understood by a participant. This process had to be done manually as synonyms of signs also counted as correct. Overall, the \textbf{average comprehension scores for all participants with both conditions was 81.56\% (\(SD = 23.29\))}.

\begin{figure}[htbp]
\centering
\includegraphics[height=4.8cm]{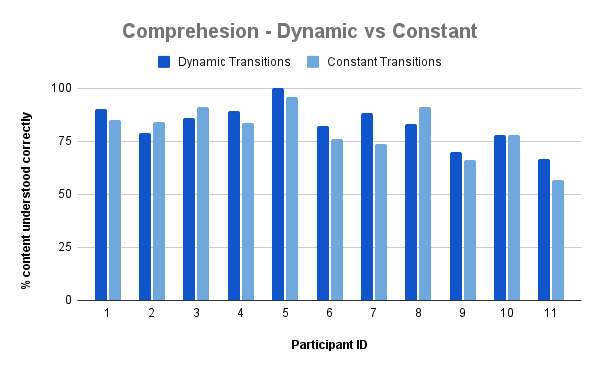}
\caption{Comprehension scores between dynamic transitions and constant transitions per participant.}
\label{graph:transitionCompr2}
\end{figure}

As shown in Figure~\ref{graph:transitionCompr2}, 7 participants had higher comprehension results in sentences with dynamic transitions, 3 participants had higher comprehension results with constant transitions and 1 participant had equal comprehension results in both transitions. According to a Shapiro-Wilk test, we retained the null hypothesis of population normality (\(p = 0.901, p = 0.722\)); therefore, we conducted a Paired samples T-test to compare differences in comprehension scores between our conditions. Based on the results, \textbf{there was no significant difference} (\(t(10) = -1.379, p = 0.198\)) in the scores for dynamic transitions (\(M = 82.97, SD = 9.43\)) and constant transitions (\(M = 80.15, SD = 11.55\)). 

In almost all cases, participants would either understand a sign or not, independently of the transition approach, which could be explained by the difference between the transition values of both approaches is not significant. However, there were 4 cases in the two-paired sentences (i.e., eight sentences) where the same sign was only perceived correctly with a dynamic approach. Moreover, there were no cases where a sign was only perceived correctly with a constant approach. Furthermore, seven participants believed transitions between signs impact comprehension, whereas only four believed they do not.

\subsubsection{Naturalness}

For each sentence in the questionnaires, we measured the percentage of naturalness by using the scores submitted on the Likert scale (i.e., one as robotic and five as natural), with 5 being 100\%. Overall, the \textbf{average naturalness scores for all participants with both conditions was 50.73\% (\(SD = 22.78\))} and the average overall naturalness\footnote{The first percentage is the average naturalness scores for all sections in the questionnaire (each video where they had to write the sentence they comprehended). The second percentage is the overall naturalness asked at the end of the questionnaire.} given at the end of the questionnaire by all participants was 50.91\% (\(SD = 25.87\)). The scores for naturalness were significantly lower than those for the other two measures, which is unsurprising because \textbf{naturalness is the most demanding criterion of all}.

\begin{figure} [htbp]
\centering
\includegraphics[height=5cm]{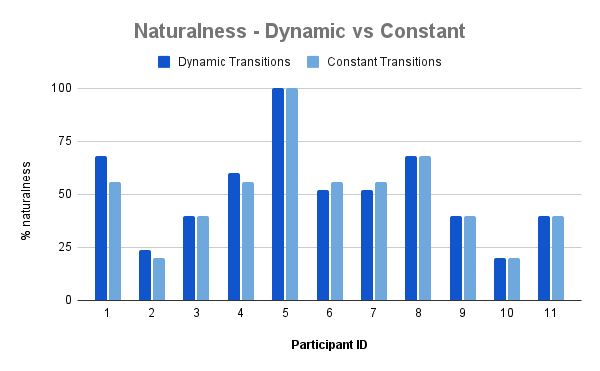}
\caption{Naturalness scores between dynamic transitions and constant transitions.}
\label{graph:transitionNatu2}
\end{figure}

As shown in Figure~\ref{graph:transitionNatu2}, there were \textbf{large discrepancies between naturalness scores throughout our participants} with 20\% as the lowest average score and 100\% as the highest score. Furthermore, 3 participants had higher naturalness results in sentences with dynamic transitions, 2 participants had higher naturalness results with constant transitions, and 6 participants had equal naturalness results in both transitions. According to a Shapiro-Wilk test, we retained the null hypothesis of population normality (\(p = 0.548, p = 0.215\)); therefore, we conducted a Paired samples T-test to compare differences in naturalness scores between our conditions. Based on the results, \textbf{there was no significant difference} (\(t(10) = -0.820, p = 0.432\)) in the scores for dynamic transitions (\(M = 51.27, SD = 22.61\)) and constant transitions (\(M = 50.18, SD = 22.51\)). However, seven participants believed transitions between signs impact naturalness, whereas only four believed they do not.

\subsubsection{Preference}

We conducted a Chi-Square test to analyze which transition approach was preferred. We found a \textbf{statistically significant relation between participants and the transition approach} (\( X^2(1, N = 33) = 6.818, p = .009 \)), as participants \textbf{preferred dynamic transitions} (\( N = 24\)) rather than constant transitions (\( N = 9\)).

\subsubsection{Transitions Speed}

For each sentence in the questionnaires, we measured the percentage of optimal transition speed by using the scores submitted in the Likert scale (i.e., one as too slow and five as too fast) with 3 being the optimal speed with 100\% and decreasing the percentage value according to the closeness to the limits of the scale with 2 and 4 as 66.67\% and 1 and 5 as 33.33\%. Overall, the average optimal transition speed scores for all participants with both conditions was 83.64\% (\(SD = 17.36\)), and the average \textbf{overall quality of transitions} given at the end of the questionnaire by all participants was 81.82\% (\(SD = 17.41\)).

\begin{figure} [htbp]
\centering
\includegraphics[height=5cm]{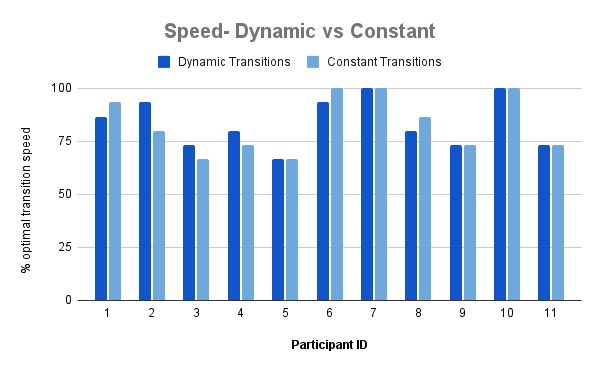}
\caption{Transition speed scores between dynamic transitions and constant transitions.}
\label{graph:transitionSpeed2}
\end{figure}

As shown in Figure~\ref{graph:transitionSpeed2}, three participants had higher optimal transition speed results in sentences with dynamic transitions, three participants had higher optimal transition speed results with constant transitions, and five participants had equal optimal transition speed results in both conditions. According to a Shapiro-Wilk test, we retained the null hypothesis of population normality (\(p = 0.283, p = 0.064\)); therefore, we conducted a Paired samples T-test to compare differences in optimal transition speed scores between our conditions. Based on the results, \textbf{there was no significant difference} (\(t(10) = -0.319, p = 0.756\)) in the scores for dynamic transitions (\(M = 83.64, SD = 11.68\)) and constant transitions (\(M = 83.032, SD = 13.45\)). However, three participants commented on the importance of faster transitions in-between signs that comprise one sole meaning and noted that constant transitions were too slow for composite utterances, and surprisingly, in negatives. The latter is one aspect we did not consider, but coincidentally our dynamic approach produced faster transitions between the negated verb and the ``NÃO'' (no) sign because the difference between hand locations of these signs is quite small. This difference allowed participants to note that constant transitions were too slow for transitions between the negated verb and the ``NÃO'' sign while our dynamic approach was optimal. We conclude that dynamic transitions might positively impact the optimal speed and that \textbf{signs that comprise one sole meaning}, as composite utterances and negation of verbs, \textbf{should have faster transitions than other signs}.

\subsubsection{Discussion}

Based on the previously reported findings, we can conclude that the null hypothesis could not be rejected in evaluating comprehension, transitions' speed, and naturalness. Nevertheless, we found particular cases where the same signs with the dynamic approach were perceived correctly and with the constant approach perceived incorrectly, but the opposite was not found. Therefore, dynamic transitions could enhance linguistic comprehension, particularly for signs that comprise one sole meaning (i.e., composite utterances and negatives) and require faster transitions. The dynamic transitions approach was also the most preferred approach by our participants, showing the positive impact they can have on animations.
  
Regarding naturalness, neither approach had a significant impact, and this criterion is still the most demanding of all. It is interesting to note that participants tend to relate naturalness to the comprehension of animations. Furthermore, it is also interesting to note that naturalness is not only linked to comprehension but also to syntax, because sentences that were completely understood but were not correct in terms of grammar, also scored lower in naturalness.

% #########################
% #############################
% #######################
\subsection{Mouthing Evaluation}\label{sec:mouthing}

We recruited 20 participants that are learning \ac{LGP}. Recruiting beginners for this user study was essential because we wanted people that had sufficient knowledge to understand some signs but not all, so that we could evaluate whether mouthing could indeed have an impact on comprehension.

Again, we recurred to a questionnaire with thirteen sentences. For this user study, we removed all phonological facial expressions from signs, so that all signs could execute mouthing. The level of complexity and difficulty in this user study was lower than the previous one, but not too easy so that we could see the impact of mouthing. This experiment was similar to the previous one (questions should evaluate naturalness, comprehension and preference). We also evaluated general quality, signs quality, and facial expressions quality in the same Likert scales. Additionally, we also asked participants whether they think ``mouthing affects naturalness", whether they think ``mouthing affects comprehension", and ``If yes, in which situations and why?".

\subsubsection{Comprehension}

We evaluated comprehension using a similar procedure to the previous study.
Overall, the \textbf{average comprehension scores for all participants with both conditions was 70.94\% (\(SD = 37.88\))} which we found surprisingly high considering that participants were beginners and sentences had a level of complexity and difficulty higher than beginner level with some sentences composed by interrogatives, one composite utterance (i.e., sign ``IRMÃ") and dactylology words comprised of numbers with two digits and names with seven letters.

\begin{figure} [htbp]
\centering
\includegraphics[height=5.1cm]{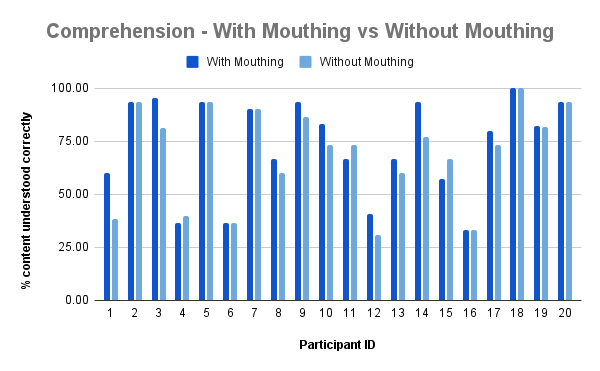}
\caption{Comprehension scores between animation with mouthing and without mouthing.}
\label{graph:mouthingCompr2}
\end{figure}

As shown in Figure~\ref{graph:mouthingCompr2}, there were large discrepancies between comprehension scores among our participants with 33.33\% as the lowest average score and 100\% as the highest score. Furthermore, 10 participants had higher comprehension results in sentences with mouthing, three participants had higher comprehension results without mouthing, and seven participants had equal comprehension results in both. According to a Shapiro-Wilk test, we rejected the null hypothesis of population normality (\(p = 0.012, p = 0.050\)); therefore, we conducted a Wilcoxon signed-rank test to compare differences in comprehension scores between our conditions. Based on these results, the \textbf{comprehension scores for sentences with mouthing were statistically significantly higher} than for sentences without mouthing (\(Z = -2.029, p = 0.043\)). Furthermore, 16 participants believe mouthing does indeed have an impact on comprehension, whereas only 4 participants believe it does not. Additionally, many comments were made by participants throughout the questionnaires noting that mouthing makes it easier to understand the sentences.

\subsubsection{Naturalness}

Naturalness was also evaluated as in the previous experiment. Overall, the \textbf{average naturalness scores for all participants with both conditions was 78.29\% (\(SD = 16.91\))} and the average overall naturalness given at the end of the questionnaire by all participants was 78.95\% (\(SD = 15.60\)). The scores for naturalness were significantly higher than scores given in the previous user study, which can be explained by the fact that participants in this study are still beginners and are still not sensible about all subtleties of sign languages; therefore, they did not notice aspects that might still be missing in our avatar.

\begin{figure} [htbp]
\centering
\includegraphics[height=5.1cm]{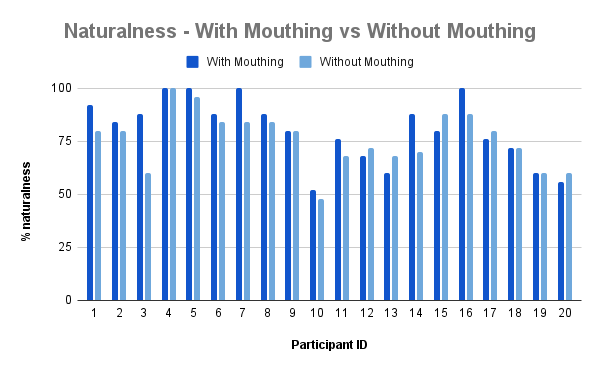}
\caption{Naturalness scores between animation with mouthing and without mouthing.}
\label{graph:mouthingNatu2}
\end{figure}

As shown in Figure~\ref{graph:mouthingNatu2}, 11 participants had higher naturalness results in sentences with mouthing, five participants had higher naturalness results without mouthing, and four participants had equal naturalness results in both. According to a Shapiro-Wilk test, we retained the null hypothesis of population normality (\(p = 0.160, p = 0.793\)); therefore, we conducted a Paired samples T-test to compare differences in naturalness scores between our conditions. Based on the results, the \textbf{naturalness scores for sentences with mouthing} (\(M = 80.40, SD = 15.24\)) \textbf{were statistically significantly higher} (\(t(19) = -2.094, p = 0.050\)) than for sentences without mouthing (\(M = 76.10, SD = 13.11\)). Furthermore, 18 participants believe mouthing has an impact on naturalness, whereas only 2 participants believe it does not.

\subsubsection{Preference}

We conducted a Chi-Square test to analyze which animations were preferred on the three trials each participant had; therefore, there were 60 trials overall. There was a \textbf{statistically significant relation between participants and the mouthing approach} (\( X^2(1, N = 60) = 15, p < 0.001 \)), as participants \textbf{preferred more animations with mouthing} (\( N = 45\)) than animations without (\( N = 15\)). One participant said that ``mouthing can distract the participant from the signs'' as being the reason for not choosing animations with mouthing.

\subsubsection{Discussion}

Based on the previously reported findings, we conclude that sentences that incorporated mouthing had higher comprehension and naturalness scores than sentences without mouthing. Therefore, our study suggests that mouthing can indeed enhance linguistic comprehension and naturalness, and participants prefer LGP animations with mouthing. 
Our user study demonstrates not only the impact mouthing has on signing animations but also that the quality of our mouthing approach was good enough to improve comprehension.

\section{Conclusions and Future Work}\label{sec:conc}

We used an existing text-to-sign language translator to demonstrate how we improved \ac{LGP} animations. We introduced a new way of performing transitions between signs -- the dynamic transitions -- and added to the translator's avatar the possibility of performing mouthing. The positive results indicate that the generated animations show great potential in the field of synthetic animation of signing avatars.

Despite the positive results, some aspects must be improved and extended in future research to improve perceived naturalness. Suggestions include adding more facial expressions, corporal movements, appropriate pauses and accelerations between signs, and creating more fluid movements.

\bibliographystyle{plainnat}
\bibliography{IVALGP}

\end{document}